\documentclass{article}
\usepackage{cite}
\usepackage{amsmath,amssymb,amsfonts}
\usepackage{algorithmic}
\usepackage{graphicx,color}
\usepackage{textcomp}
\usepackage{xcolor}
\usepackage{hyperref}
\usepackage{pgfplots}
\usepackage{graphicx}
\usepackage{amsmath, amssymb}
\usepackage{todonotes}
\usepackage{subcaption}
\usepackage{titlesec}
\hypersetup{hidelinks=true}
\usepackage{algorithm,algorithmic}
\newtheorem{definition}{Definition}[section]

\usepackage[final,nonatbib]{neurips_2024}

\usepackage[utf8]{inputenc} % allow utf-8 input
\usepackage[T1]{fontenc}    % use 8-bit T1 fonts
\usepackage{hyperref}       % hyperlinks
\usepackage{url}            % simple URL typesetting
\usepackage{booktabs}       % professional-quality tables
\usepackage{amsfonts}       % blackboard math symbols
\usepackage{nicefrac}       % compact symbols for 1/2, etc.
\usepackage{microtype}      % microtypography
\usepackage{xcolor}         % colors

\title{Knowledge Distillation-Based Model Extraction Attack using GAN-based Private Counterfactual Explanations}

\author{
  Fatima Ezzeddine \\
  Faculty of Informatics \\
  Università della Svizzera italiana\\
  \texttt{fatima.ezzeddine@usi.ch} \\
  \And
  Omran Ayoub \\
  Department of Innovative Technologies \\
  University of Applied Sciences and \\ Arts of Southern Switzerland \\
  \texttt{omran.ayoub@supsi.ch} \\
  \AND
  Silvia Giordnao \\
  Department of Innovative Technologies  \\
  University of Applied Sciences and Arts \\ of Southern Switzerland  \\
  \texttt{silvia.giordano@supsi.ch} \\
}

\begin{document}

\maketitle

\begin{abstract}
In recent years, there has been a notable increase in the deployment of machine learning (ML) models as services (MLaaS) across diverse production software applications.
In parallel, explainable AI (XAI) continues to evolve, addressing the necessity for transparency and trustworthiness in ML models. XAI techniques aim to enhance the transparency of ML models by providing insights, in terms of \emph{model's explanations}, into their decision-making process. Simultaneously, some MLaaS platforms now offer explanations alongside the ML prediction outputs. This setup has elevated concerns regarding vulnerabilities in MLaaS, particularly in relation to privacy leakage attacks such as model extraction attacks (MEA). This is due to the fact that explanations can unveil insights about the inner workings of the model which could be exploited by malicious users. In this work, we focus on investigating how model explanations, particularly counterfactual explanations (CFs), can be exploited for performing MEA within the MLaaS platform. We also delve into assessing the effectiveness of incorporating differential privacy (DP) as a mitigation strategy. To this end, we first propose a novel approach for MEA based on Knowledge Distillation (KD) to enhance the efficiency of extracting a substitute model of a target model exploiting CFs, without any knowledge about the training data distribution by the attacker. Then, we advise an approach for training CF generators incorporating DP to generate private CFs. We conduct thorough experimental evaluations on real-world datasets and demonstrate that our proposed KD-based MEA can yield a high-fidelity substitute model with a reduced number of queries with respect to baseline approaches. Furthermore, our findings reveal that including a privacy layer can allow mitigating the MEA. However, on the account of the quality of CFs, impacts the performance of the explanations.
\end{abstract}

\section{Introduction}
Recent years have witnessed a growing trend in employing deep neural networks (DNNs) as learning algorithms for machine learning-based applications. In particular, DNNs have gained substantial popularity due to their remarkable success in diverse domains \cite{shinde2018review}. However, the complexity of their training process, which is resource-intensive, involves the acquisition of data and requires substantial computational power \cite{lecun2015deep}, hinders their widespread adoption and accessibility. An approach to addressing this challenge is by offering machine learning (ML) models as a service through dedicated ML as a Service (MLaaS) platforms \cite{nasr2019comprehensive, shokri2021privacy}. MLaaS platforms allow to streamline the accessibility of such complex models by hosting and training ML models on the cloud, allowing third-party practitioners to access these pre-trained models through Application Programming Interfaces (APIs) \cite{wang2022dualcf}, while the private dataset used to train the model is inaccessible. %This method not only facilitates the developer's workload but also makes advanced ML more accessible to everyone.

Numerous platforms offer MLaaS such as Google Cloud, Microsoft, IBM, and Amazon Web Services \cite{AzureAI, googleAutoML, MicrosoftXAI, IBMXAI, googleXAI, AWSXAI}. The API's input-output format for these services is publicly accessible, meaning that users have knowledge of the format of input data required by the service (by the ML model), and can interpret its outputs. This opens the door for malicious users (or, more precisely, attackers) to attempt to perform privacy-breaching attacks such as membership inference attacks (MIA) \cite{shokri2017membership}, model inversion attacks (MINA) \cite{wang2021variational}, and model extraction attacks (MEA) \cite{tramer2016stealing}. For instance, in MIA, an attacker aims to infer the presence of a specific individual instance in the private training set. In MINA, the attacker attempts to reconstruct the training set. In MEA, the attacker attempts to extract the model by training a substitute model using data acquired through repetitive queries to the service provider's confidential model. Several mitigation strategies for these security- and privacy-breaching attacks are already employed \cite{dwork2006calibrating}. However, these strategies are faced with new challenges as emerging transparency requirements demand explanations for ML model decisions. In fact, MLaaS platforms are shifting towards providing explanations alongside the predictions made by the deployed ML models \cite{MicrosoftXAI, googleXAI, AWSXAI, IBMXAI}. The need to provide users with such model explanations is linked to providing users with a transparent decision-making process of data-driven systems \cite{nguyen2024survey}. The urge to enhance transparency through model explanations aligns, on one hand, with the aim of meeting recent regulatory (and compliance) demands, and on the other hand, with the aim of providing users with actionable insights \cite{nguyen2023feasible,thang2015evaluation,nguyen2024survey}. Such explanations are extracted using explainable AI (XAI) techniques, which aim to provide insight into how a model arrived at its decision \cite{holzinger2019causability, ribeiro2016should}. In this context, explanations may inadvertently provide attackers with insights to enhance their attacks, posing a fresh obstacle to existing security and privacy measures \cite{huang2024explaining,huangaccurate,pawelczyk2023privacy}.

In this work, we focus on investigating how example-based explanations, specifically counterfactual (CF) explanations, can be exploited to perform MEA within MLaaS scenarios. We precisely focus on CFs due to the unique insights they provide, as they reveal how to minimally alter original data instances to achieve a different model's outcome, allowing users to interpret the dynamics underlying the model's prediction shift from the decision boundary \cite{dovsilovic2018explainable}. However, due to the proximity of CFs to the decision boundary and their representativeness of the training data, CFs possess a dual nature, serving not only as interpretative tools but also as potential assets for enhancing attacks as they provide additional insights and knowledge for potentially exploiting the vulnerabilities of the model under attack. Specifically, we propose a novel approach based on the concept of knowledge distillation (KD) to perform high-fidelity MEA with CFs as the representation of data, and derive a threat model that have similar functionality of the original model being targeted \footnote{Link to code: https://github.com/FatimaEzzedinee/Knowledge-Distillation-Based-Model-Extraction-Attack-using-GAN-based-Private-CFs}. The unique capabilities of KD allow us to handle the problem not just as a standard classification task but as an estimation of the output probability distribution. Moreover, we propose a mitigation approach that employs differential privacy (DP) within the CF generator training pipeline, aiming to reduce the risk associated with providing CF explanations. Our analysis examines the attack performance of our proposed method compared to baselines. Additionally, we assess the impact of DP on the quality of CF and the performance of the MEA. The contribution of this paper is summarized as follows:
\vspace{-3pt}
\begin{itemize}
    \item \textbf{KD-Based MEA with CFs}: This research investigates potential vulnerabilities in MLaaS, with a specific focus on MEA facilitated by KD as extraction techniques and CFs as representatives of the training set. The research simulates an adversarial scenario wherein an attacker employs our proposed KD techniques to extract a substitute for a target model.

   \item \textbf{Private Counterfactual Explanations}: Recognizing the importance of preserving the privacy of training data, the paper introduces the concept of DP within the GAN-generated CF explanation pipeline. This contribution aims to generate CFs that deviate from the statistical properties of the confidential dataset, offering a layer of protection against potential privacy breaches.
\end{itemize}

\section{Related work}
\vspace{-2pt}
Several works have proposed strategies for performing MEA for classification. Authors in \cite{tramer2016stealing} perform successful MEA on different ML models like decision trees, Support Vector Machines, and DNNs by using equation-solving and path-finding algorithms and learning theory. In \cite{pal2020activethief} authors perform MEA on DNNs using active learning with unannotated public data. In \cite{krishna2019thieves}, authors perform MEA in natural language processing on bert-based APIs, in which they explore transfer learning for MEA. Other works have focused on proposing querying strategies for MLaaS to effectively query the target model and extract accurate insights in addition to performing MEA \cite{orekondy2019knockoff, juuti2019prada, oh2019towards}. In \cite{orekondy2019knockoff} authors train a knockoff network with queried image predictions and propose a reinforcement learning approach that additionally improves query sample efficiency in certain settings and provides performance gains. In \cite{juuti2019prada} authors propose a new method that generates synthetic queries and optimizes training hyperparameters, and then propose a method to detect generic and effective detection of DNN MEA. 
%They analyze the distribution of consecutive API queries and raise an alarm when this distribution deviates from normal behavior. 
In \cite{oh2019towards} authors investigate the type and amount of internal information about the black-box model that can be extracted from querying the deployed model, such as the architecture, optimization procedure, or training data. Authors in \cite{jagielski2020high} improve the query efficiency of attacks by designing learning-based methods and focusing on the real accuracy of the extracted model. Also \cite{krishna2023towards}, discuss the tension between the right to explanation
and the right to be forgotten for privacy concerns.

Other works have investigated how to reveal insights of ML models using the explanations provided by the MLaaS \cite{aivodji2020model, oksuz2023autolycus, yan2022towards, sokol2019counterfactual, milli2019model}. In \cite{oksuz2023autolycus} authors investigate how Local Interpretable Model-agnostic Explanations (LIME), an XAI framework, can be exploited to infer the decision boundaries by sending adaptive queries to generate new data samples that are close to the decision boundaries. In \cite{miura2021megex} authors analyze how gradient-based explanations reveal the decision boundary of a target model by modeling a data-free MEA against a gradient-based XAI, in addition to exploiting a generative model to reduce the number of queries. \cite{yan2022towards} proposed a methodology to perform MEA by minimizing task‐classification loss and task‐explanation loss. Similar to our work, other efforts have investigated how CFs can be exploited in MEA. In \cite{aivodji2020model} authors model an attack that relies on both the predictions and the CFs of the target model to directly train an attack model. The authors in \cite{wang2022dualcf} model a strategy to reveal the decision boundary of a target model and then perform MEA by considering the CF of the CF as pairs of training samples to directly train the extracted model.

Similar to these works, our work aligns with exploring how CF explanations can be exploited for performing MEA. However, our contribution lies in modeling a novel approach to MEAs that exploits more effectively the insights that CFs carry with respect to other approaches that directly train a substitute model on extracted CFs. Moreover, in contrast to previous studies \cite{aivodji2020model, wang2022dualcf}, our work presumes that the attacker possesses no prior knowledge of the training set, leading them to query the model with zero knowledge, and propose a novel approach for performing the MEA.

% We propose a novel methodology to perform MEA that employs KD. We consider the approach proposed in \cite{aivodji2020model} as a baseline approach with respect to ours. 

In terms of mitigation strategies, few works have focused on proposing methodologies to generate explanations while mitigating the revelation of sensitive insights about the decision boundary, the training set, or the model architectures. For instance, authors in \cite{patel2022model} propose employing DP algorithms to construct feature-based model explanations, while authors in \cite{yang2022differentially} propose an approach to generate differentially private CFs via functional mechanisms. \cite{pentyala2023privacy} proposes an approach for generating recourse paths leveraging differentially private clustering and demonstrates that constructing a graph on the cluster centers provides private recourse paths as CFs.

Also, our work also exploits DP, however, by integrating it into the CF generation process, and eliminates the necessity for a distinct method to generate private CFs. We leverage the strength of generative adversarial networks (GANs)-based generators in producing high-quality, yet private, CFs. To the best of our knowledge, our work is the first to adapt DP within CF generative-based methods. We specifically focus on GANs-based CF generators due to their ability to generate high-quality CFs compared to other CF generation methods. We show the robustness of the DP-based proposed method against MEA, and we provide a detailed analysis that sheds light on the impact of CFs in exposing the distribution of the training set through well-defined metrics.

\section{Scenario and problem formulation}
A DNN model $f_{\theta}$ trained on a dataset $\mathcal{D}:$ $\mathcal{X} \in \mathbb{R}^d \rightarrow \mathcal{Y} \in \mathbb{R}$ takes a numerical input query $x \in \mathcal{X}$ and predicts the output $y \in \mathcal{Y}$. $f_{\theta}$ a trained model with $\theta$ be the weight matrix, $b$ be the bias vector. The output of the DNN can be computed as $f_{\theta} = \theta \cdot X + b$. The output of the network $y$ are confidence scores, which are expressed in terms of probabilities, essentially indicating how confident the ML model is about each possible class or category in its training data. The CF explanation method, which trains a CounterGAN generator $E$: $f_{\theta} \times \mathcal{X} \rightarrow \mathcal{X}$ generates a perturbed instance $c \in \mathcal{X}$ for the input instance $x$ such that $f_{\theta}(c)$ has a different output while minimizing an objective function $g$. Formally, searching for a CF can be framed as $argmin$ $g(x, c)$ $s.t$ $f_{\theta}(c) \neq f_{\theta}(x)$. Where $g: \mathcal{X} \times \mathcal{X} \rightarrow \mathcal{R}_{+}$ is a cost metric measuring the changes between the input $x$ and $c$, and $y$ is the desirable target that is different from the original prediction $f_{\theta}$.% That is, we seek to find counterfactual explanations that belong to a target class $y$ while remaining proximal to the original instance.

%We consider a scenario where an MLaaS API provides predictions for a specific ML task along with the corresponding explanations, in terms of CFs, as insight into the AI model's decision-making process. The API takes an instance query denoted as $x$ as input, evaluates the input using a model represented by $f_{\theta}(x)$, and returns the prediction $y$ and its corresponding CF explanation $E (x)$.

\begin{figure}[ht]
\centering
\includegraphics[scale=0.3]{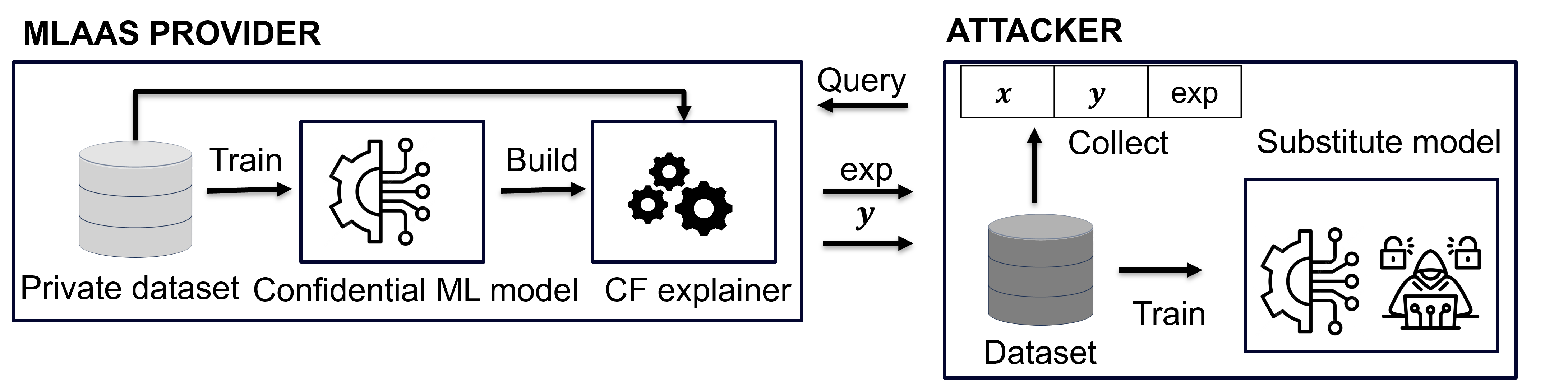}
\caption{Scenario of MEA where MLaaS provides explanations alongside the prediction}
\label{scenario}
\end{figure}

Figure \ref{scenario} depicts the steps of the scenario. First, a user sends a query, which includes input data describing a data record $x$ for which the user aims to receive the prediction. Once the MLaaS API receives the user query, it will pass it to the ML service, which performs the prediction $f_{\theta}(x)$ and generates a CF explanation $c=E(x)$ where $f_{\theta}(c)$ has a different prediction and finally returns it along with its corresponding output to the user.

Within the scope of this research, our primary emphasis is placed on a specific type of attack known as the high-fidelity MEA, specifically directed toward CFs. The objective of this attack is to derive a model threat\_model $t_{\Upsilon}$ that closely resembles, in terms of functionality, the original model being targeted.% This derived model is often referred to as a threat\_model $t_{\Upsilon}$.
%Note that this research delves into the privacy implications of CF explanations; hence, the goal is not just to replicate the target model but to prove the issues in exposing CFs.
The MEA is formulated as follows: for a set of queries $\mathcal{Q}$ and a set of corresponding CF explanations $\mathcal{C}$, an attacker trains a threat model $t_{\Upsilon}$ that performs equivalently on an evaluation set $\mathcal{T}$. The goal is to maximize an agreement function between $f_{\theta}$ and $t_{\Upsilon}$ as shown in Eq. \ref{agreement}, with a minimal number of queries sent to the API. The agreement metric will be detailed later in this paper in Sec. \ref{EvaluationSettings}.
\vspace{-15pt}
%In essence, the central theme of this paper revolves around the concept of high-fidelity extraction attacks, with a specific interest in their application to CF, and the ultimate goal is to produce a model that behaves in a manner closely resembling the original model, hence referred to as a "threat model"
%In this paper, we focus on the high-fidelity extraction attack with CFs, which aims to extract a functional equivalent or near-equivalent model called a threat model $t_{\Upsilon}$ that behaves very similarly to the model $f_{\theta}$.
\section{Methodology}\label{methodolgySec}
%In this section, we first describe our proposed KD-based MEA and train a threat model $t_{\Upsilon}$. Then, we describe our proposed methods for integrating DP within the CF generation process.
\subsection{Knowledge distillation-based model extraction attack}
\begin{figure}[]
\centering
\includegraphics[scale=0.4]{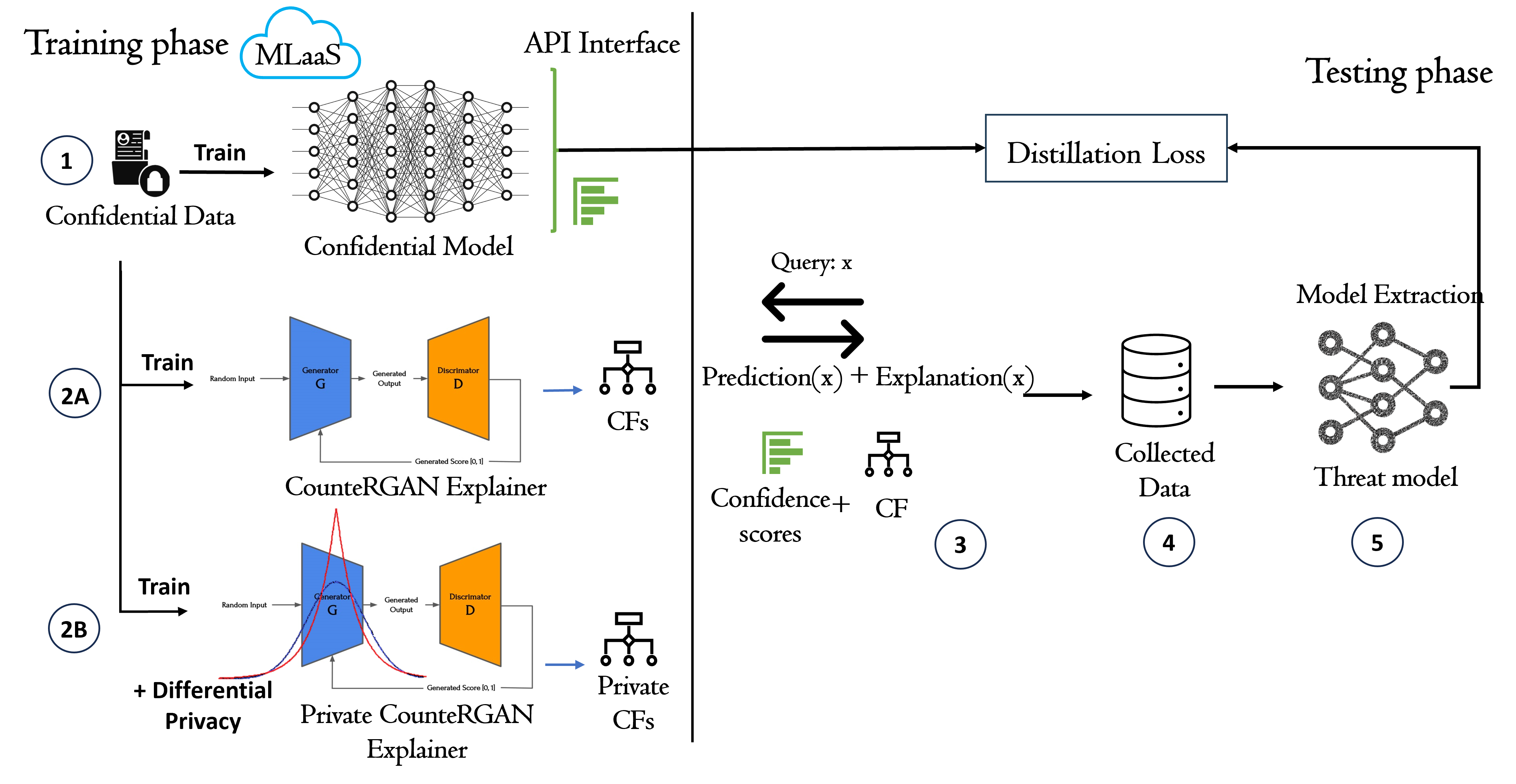}
\caption{Methodology Framework, describing the KD-based MEA, in the deployment phase, in addition to the CF generation phase.}
\label{methodolgy}
\end{figure}

First, an owner of a private dataset trains a DNN classifier (Step 1 in Fig. \ref{methodolgy}), and a CounterGAN CF explainer (Step 2A, 2B in Fig. \ref{methodolgy}) and deploys it on an MLaaS platform.
As an attacker, specifically, we query the model with random queries \footnote{The attacker does not have previous knowledge of the training set.} and collect the query output of the target model, and the CFs explaining the output of the target model (Step 3 in Fig. \ref{methodolgy}). After successfully collecting a set of CFs (Step 4 in Fig. \ref{methodolgy}), the attacker employs our KD-based MEA approach on the collected dataset as input data to train $t_{\Upsilon}$. The rationale for selecting KD as the extraction method lies in its unique capability to address the problem not only as a standard supervised classification task but also in its proficiency in precisely imitating the probability distribution. Specifically, we propose to train $t_{\Upsilon}$ by minimizing the loss constituted by the threat model classification loss in addition to the distillation loss. To emphasize the importance of mimicking the output probabilistic distribution from the target model to the threat model, we set the distillation loss to the Jensen-Shannon (JS) (\ref{JS}) metric. The choice of JS is to tackle issues related to divergence and ensure a symmetric output, so we suppose that the attacker replaces the KL divergence metric with JS. Afterward, the attacker trains the threat model by minimizing the loss of KD (Step 5 in Fig. \ref{methodolgy}) until the convergence of the agreement on a separate validation set.
\begin{equation}
\label{JS}
    \text{JS}(P\|Q) = \frac{1}{2}{\text{KL}}\left(P\left\|\frac{P+Q}{2}\right.\right) + \frac{1}{2}{\text{KL}}\left(Q\left\|\frac{P+Q}{2}\right.\right)
\end{equation}
\subsection{Private CFs generation with DP}
We are now aligned with the service provider and are presenting a strategy to mitigate MEA by avoiding providing Private CFs. Our goal is to examine the effects of incorporating DP into the CF generation process on MEA performance and the quality of the provided CFs considering metrics such as actionability, prediction gain, and realism, which are explained later in sec \ref{EvaluationMetrics}. Additionally, we aim to demonstrate how private CFs align with the distribution of the training set. We propose an approach to integrating DP into the CF generation process and generating private CFs. Our methodology does not memorize or expose statistical information about the training set. We propose to incorporate DP into the generator of the CounteRGAN. The primary objective is to prevent the generation of CFs that closely resemble the private training data, thereby reducing the resemblance of CFs with the training data.

To this end, we inject DP into the generator during the optimization process. More specifically, we employ the Adam differential private optimizer (DP Adam). The process of DP Adam often involves multiple iterations of the training process to achieve privacy guarantees. We add noise repeatedly over several rounds, so the overall privacy guarantees are strengthened. To this end, we inject noise into the gradients of the generator in addition to a step of gradient clipping. The process involves clipping the gradient for a random subset of examples, clipping the norm of each gradient, computing the average, adding calibrated noise, and finally performing the traditional back-propagation step that updates the weights. Note that the gradient clipping step is essential as it controls the amount of noise added and prevents large updates that could reveal too much information about individual data points. The key assurance provided by this approach is that the generator ensures that it does not memorize or reveal details of the training data and therefore does not generate CFs that are very similar to the training points.

%By embedding DP directly into the generator's loss function and leveraging a DP optimizer, we aim to establish a robust safeguard against the leakage of sensitive information during the generation of CFs while maintaining the quality of CFs.

%Injecting DP into the CF generator offers a more balanced approach, allowing for improved privacy while minimizing potential impacts on the overall performance of the model. This strategic choice acknowledges the need to safeguard sensitive information in the generated CFs while maintaining an optimal balance between model accuracy and privacy preservation within the explanations.
\section{Experimental settings} \label{EvaluationSettings}

\subsection{Baseline scenarios}
We compare our proposed approach for performing MEA, KD-based MEA, to a baseline approach, presented in \cite{aivodji2020model}, and referred to as \emph{Direct Train}, which trains a threat model as a standard supervised classification task. 

We examine three scenarios pertaining to the MLaaS provider's inclusion or exclusion of CF explanations alongside the predictions of the ML model, 1) \emph{No CF:} In this scenario, the MLaaS does not provide the user with any CFs. 2) \emph{CF:} In this case, the MLaaS provides CFs suing counterGAN. \emph{3) Private CF:} The MLaaS offers private CFs by implementing our proposed approach for generating private CFs, discussed in Sec. \ref{methodolgySec}.

Applying each MEA approach to the three scenarios results in the following six scenarios: 1) \emph{KD-based Private CF.} We employ our proposed KD-based approach on CFs extracted from the differentially private generator, and the focus shifts to examining the potential effect of integrating DP in the CF generation. 2) \emph{KD-based CF.} We employ our proposed KD-based approach, trained with CFs, to understand the influence of KD when applied alongside CF. 3) \emph{KD-based No CF.} We employ our proposed KD-based approach on the randomly generated data points, and the focus shifts to examining the potential enhancement when KD is utilized independently of CFs and analyzing the power of CFs alongside. 4) \emph{Direct No CF.} The MEA is performed following the baseline \cite{aivodji2020model}. We train a model directly on the randomly generated data points, and the focus shifts to examining the potential of direct queries and analyzing the power of KD and CFs. 5) \emph{Direct CF.} We train a model directly on CFs, this dimension is dedicated to evaluating the standalone power of CFs without the incorporation of KD, providing insights into the capabilities of CFs separately. 6) \emph{Direct Private CF.} We train a model directly on CFs extracted from the differentially private generator, the focus shifts to examining the potential of direct queries and to analyzing the power of KD and CFs alongside.

%\section{Models Training}
\subsection{Target model}
We adhere to consistent training procedures for the target model (to be deployed and queried by the API). We train a DNN comprising of 16 hidden layers to create a complex model. The layer configurations consist of 64, 32, 16, 32, 64, 128, 64, 32, 128, 64, 128, 64, 128, 64, 32, and 16 neurons in each layer, respectively. The activation function is set to gelu \cite{hendrycks2016gaussian} for all layers, while the output layer employs softmax to produce model outputs as confidence score probabilities. We specify the optimizer as the Adam optimizer to minimize the cross-entropy loss function, with default parameter initialization provided by Keras. Each target model is trained using 80\% of the training dataset \footnote{Experiments were computed with a machine of intel Core i7, a GPU of RTX 3070, and 8 GB of RAM.}.
\subsection{Threat model}
We specify consistent training procedures for the threat model, with the assumption that the attackers do not know the target model architecture but can build a DNN with a standard architecture. The attacker aim to train either by using our proposed approach of KD-based MEA or directly on the data points a DNN with 3 hidden layers of 16, 32, and 64 neurons consecutively, with an activation function as the relu, and an output layer with a softmax activation function. In the scenario we are considering, we assume that the attacker does not possess any knowledge about the distribution of the training data used for training the target model. Consequently, the attacker generates 1000 random data points for each dataset. The data point values are randomly generated, specifying the values of a range between -3 and 3 for each feature. We specify -3 and 3 as ranges for random values to have as randomized as possible data points that are not similar to the training datasets.
For the training with KD-based MEA, we vary $\alpha$ between 0 and 0.5 and report the highest agreement. We made sure that the results were reproducible by fixing the seed and the layer initialization with \emph{LeCun Initializer}. We report the results of the average of 10 runs with randomly chosen subsets for each experiment.
\subsection{Counterfactual generator}
To generate CFs using CountRGAN, we train a discriminator structured with three layers comprising 32, 16, and 1 neurons consecutively. Each hidden layer is followed by a dropout layer with a factor of 0.2. We use a relu activation function for the hidden layers and a sigmoid activation function for the output layer. The generator, which takes a datapoint as input and produces the corresponding CF, is trained with three hidden layers comprising 64, 48, and 32 neurons, using relu activation for the hidden layers and a linear activation function for the output layer. We specify the optimizer as Adam. For the DP generator, we change the optimizer to be an optimizer that supports DP. We integrate the DP Adam optimizer from the TensorFlow privacy library. To guarantee a good privacy budget, we specify the parameters of the DP Adam optimizer as l2\_norm\_clip to 1, the noise\_multiplier to 3.%, and num\_mictobatches to 1.

\subsection{Evaluation metrics}\label{EvaluationMetrics}
The evaluation metrics comprise two sets. The first set measures the effectiveness of MEA while the second encompasses a pair of metrics to quantify the quality of the CFs.

\textbf{Effectiveness of MEA}: A commonly employed metric for assessing the effectiveness of MEA is the \emph{Agreement} measure. \emph{Agreement} measures the degree of alignment between two models, i.e., the similarity in the predictions between the two different ML models. In the context of MEA methods, agreements assess the output similarity between the predictions of an extracted model $t_{\Upsilon}$ and those of the target model $f_{\theta}$ for a given set of data records (see Eqn. \ref{agreement}). Since our goal is to replicate the behavior of a target model with an extracted model, a higher agreement means a more successful MEA, and hence, a more effective strategy for MEAs.
\begin{equation}
    \label{agreement}
     agreement(f_{\theta}, t_{\Upsilon}) = \sum_{x_i \in \mathcal{T}} {1}_{f_{\theta} = t_{\Upsilon}}
\end{equation}
\textbf{Success of explainer and quality of CFs}: To measure the quality of extracted CFs, and to analyze the explainer prediction classes shift we employ three commonly used metrics, namely, \emph{Prediction Gain}, \emph{Realism}, and \emph{Actionability}.

The \emph{Prediction Gain} quantifies how much the probability of the CF explanation for a specific target class $t$ changes in comparison to the original data point as provided by the classifier for a given target class $t$. In other words, it allows us to quantify to what extent the explainer was capable of shifting the prediction of the classifier using the extracted CFs. In our case, since the classifier's score is in terms of probabilities, the prediction gain spans from 0 to 1, where a higher prediction gain indicates that the explainer shows a stronger shift towards a target class $t$. 

The \emph{Realism} is a metric used to quantify how closely a data instance fits a known data distribution. Since in our work, we employ DP to avoid providing CF that may expose statistical information about the training set, we employ realism to quantify how well CFs and private CFs fit in with the known data distribution. Furthermore, realism is employed to compare a data point with its associated CF, to assess the authenticity of CFs and to examine the impact of the data perturbation with DP on CF. Inspired by the strategies proposed in \cite{nemirovsky2020countergan, joshi2019towards}, we train a denoising autoencoder on the noised training set, and compute the realism of a data point as the reconstruction error as the mean squared error (MSE) of the autoencoder (Eq. \ref{reconstructionError}). A lower value of $realism$ corresponds to a better fit of the data point within the data distribution.
\begin{equation}
\text{Realism} = \frac{1}{N} \sum_{i=1}^{N} \| \text{input}_i - \text{reconstruction}_i \|^2
\label{reconstructionError}
\end{equation}

\emph{Actionability} is a commonly used metric that allows to measure the quality of a CF. \emph{Actionability} assumes both the number of altered features and the degree of those changes (amount of modification in a feature) in a CF relative to the original data instance. We compute \emph{Actionability} by taking the L1 norm of the absolute difference between the data point and its corresponding CF (Eq. \ref{actionability}. A low value of actionability (desired) indicates that a relatively small subset of the input features has been perturbed, with respect to the original data instance, to achieve a different outcome by the classifier, and hence, a low \emph{actionability} suggests better quality CFs.
\begin{equation}
\text{Actionability} = \frac{1}{N} \sum_{i=1}^{N} \| \text{input}_i - \text{CF}_i \|^1
\label{actionability}
\end{equation}
By computing this comprehensive set of metrics, we aim to explore how DP impacts the feasibility of taking actionable steps and how this influence contributes to the prediction gain outcomes.% Additionally, we aim to understand how these effects reflect on the realism of CFs.
%\vspace{-4pt}
\section{Results and evaluation}
%Agreement for Model Extraction Attack Results}
\begin{figure*}[t]
  \centering
  \hspace{-60pt}
  \begin{subfigure}[b]{0.27\textwidth}
    \begin{tikzpicture}
        \begin{axis}[
            xmode = log,
            width=165pt,
            height=120pt,
            xlabel={Queries Number},
            ylabel={Agreement},
            xmin=45, xmax=1050,
            ymin=55, ymax=95,
            xtick={50, 100, 200, 300, 500, 1000},%, 2000, 3000, 4000}, % Added ticks at 750 and 1000
            xticklabels={50, 100, 200, 300, 500, 1000},%, 2000, 3000, 4000}, % Labels for the ticks
            ytick={60, 70, 80, 90, 100},
            %legend pos=outer north east,
            %legend style = {draw=none},
            legend style={at={(0.3,1.05)}, anchor=south, legend columns=1, draw=none},
            ymajorgrids=true,
            grid style=dashed,
            y label style={at={(axis description cs:0.2,.5)},anchor=south},
            ]
            \addplot[
                color=red,
                mark=square,
                ]
                coordinates {
                (0, 0)(50, 83)(100, 85)(200,87)(300, 88)(500,90)(1000,91)%(2000, 0)(3000, 0)(4000, 0)
                };
                \addlegendentry{\small{KD-based CF}}
            
            \addplot[
                color=blue,
                mark=square,
                ]
                coordinates {
                (0, 0)(50, 80)(100, 82)(200,83)(300, 83)(500,84)(1000,86)%(2000, 0)(3000, 0)(4000, 0)
                };
                \addlegendentry{\small{Direct CF}}
            
                \addplot[
                color=black,
                mark=square,
                ]
                coordinates {
                (0, 0)(50, 57)(100, 68)(200,74)(300, 75)(500,78)(1000,79)%(2000,0)(3000, 0)(4000, 0)
                %(0, 0)(50, 64)(100, 74)(200,77)(300, 77)(500,0)(1000,0)(2000, 0)(3000, 0)(4000, 0)
                };
                %\addlegendentry{KD - no CF}
             
            \addplot[
                color=green,
                mark=square,
                ]
                coordinates {
                (0, 0)(50, 57)(100, 64)(200,71)(300, 73)(500,75)(1000,76)%(2000, 0)(3000, 0)(4000, 0)
                %(0, 0)(50, 63)(100, 73)(200,76)(300, 0)(500,0)(1000,0)(2000, 0)(3000, 0)(4000, 0)
                };
                %\addlegendentry{Direct - no CF}
            
                \addplot[
                color=violet,
                mark=triangle,
                ]
                coordinates {
                (0, 0)(50, 60)(100, 68)(200,76)(300, 77)(500, 81)(1000,83)%(2000, 0)(3000, 0)(4000, 0)
                };
                %\addlegendentry{KD - DP CF}
    
               \addplot[
                color=teal,
                mark=triangle,
                ]
                coordinates {
                 (0, 0)(50, 60)(100, 67)(200,75)(300, 76)(500, 79)(1000,81)%(2000, 0)(3000, 0)(4000, 0)
                };
                %\addlegendentry{Direct - DP CF}
                %\legend{}; 
        \end{axis}
    \end{tikzpicture}
    \caption{GMSC Dataset}
    \label{fig:gmcs}
   \end{subfigure}
   \hspace{40pt}
    \begin{subfigure}[b]{0.27\textwidth}
        \begin{tikzpicture}
        \label{CreditFraudDatasetAgreement}
            \begin{axis}[
                xmode = log,
                width=165pt,
                height=120pt,
                xlabel={Queries Number},
                xmin=45, xmax=1050,
                ymin=50, ymax=95,
                xtick={50, 100, 200, 300, 500, 1000},%, 2000, 3000, 4000}, % Added ticks at 750 and 1000
                xticklabels={50, 100, 200, 300, 500, 1000},%, 2000, 3000, 4000}, % Labels for the ticks
                ytick={60, 70, 80, 90, 100},
                legend style={at={(0.3,1.05)}, anchor=south, legend columns=1, draw=none},
                %legend pos=outer north east,
                %legend style = {draw=none},
                ymajorgrids=true,
                grid style=dashed,
                y label style={at={(axis description cs:0.2,.5)},anchor=south},
                ]
                
            \addplot[
                color=black,
                mark=square,
                ]
                coordinates {
                 (0, 0)(50, 60)(100,65)(200,67)(300, 71)(500,73)(1000,81)%(2000,0)(3000, 0)(4000, 0)
                %(0, 0)(50, 64)(100, 74)(200,77)(300, 77)(500,0)(1000,0)(2000, 0)(3000, 0)(4000, 0)
                };
                \addlegendentry{\small{KD-based No CF}}
             
            \addplot[
                color=green,
                mark=square,
                ]
                coordinates {
                 (0, 0)(50, 58)(100,63)(200,65)(300,67)(500, 70)(1000,75)%(2000, 0)(3000, 0)(4000, 0)
                %(0, 0)(50, 63)(100, 73)(200,76)(300, 0)(500,0)(1000,0)(2000, 0)(3000, 0)(4000, 0)
                };
                \addlegendentry{\small{Direct No CF}}
                
            \addplot[
                color=red,
                mark=square,
                ]
                coordinates {
                (0, 0)(50, 87)(100, 90)(200,91)(300, 92)(500, 92)(1000,93)%(2000, 0)(3000, 0)(4000, 0)
                };
                %\addlegendentry{KD - CF}

            \addplot[
                color=blue,
                mark=square,
                ]
                coordinates {
                 (0, 0)(50, 83)(100, 86)(200,88)(300, 89)(500,89)(1000,90)%(2000, 0)(3000, 0)(4000, 0)
                };
                %\addlegendentry{Direct CF}

            \addplot[
                color=violet,
                mark=triangle,
                ]
                coordinates {
                (0, 0)(50, 63)(100, 64)(200,68)(300, 69)(500, 77)(1000,82)%(2000, 0)(3000, 0)(4000, 0)
                };
                %\addlegendentry{KD - DP CF}
            
            \addplot[
                color=teal,
                mark=triangle,
                ]
                coordinates {
                 (0, 0)(50, 52)(100, 58)(200,63)(300, 67)(500, 74)(1000,79)%(2000, 0)(3000, 0)(4000, 0)
                };
                %\addlegendentry{Direct - DP CF}
            %\legend{}; 
            \end{axis}
        \end{tikzpicture}
        \caption{Credit Card Fraud Dataset}
        \label{fig:credit_fraud}
       \end{subfigure}
    \hspace{30pt}
      \begin{subfigure}[b]{0.27\textwidth}
        \begin{tikzpicture}
        \begin{axis}[
            xmode = log,
            width=165pt,
            height=120pt,
            xlabel={Queries Number},
            xmin=45, xmax=1050,
            ymin=55, ymax=83,
            xtick={50, 100, 200, 300, 500, 1000},%, 2000, 3000, 4000}, % Added ticks at 750 and 1000
            xticklabels={50, 100, 200, 300, 500, 1000},%, 2000, 3000, 4000}, % Labels for the ticks
            ytick={60, 70, 80, 90, 100},
            %legend pos=outer north east,
            ymajorgrids=true,
            grid style=dashed,
            legend style={at={(0.3,1.05)}, anchor=south, legend columns=1, draw=none},
            y label style={at={(axis description cs:0.2,.5)},anchor=south},
            ]
                                    
        \addplot[
            color=violet,
            mark=triangle,
            ]
            coordinates {
            (0, 0)(50, 59)(100, 60)(200,62)(300, 65)(500, 67)(1000,68)%(2000, 0)(3000, 0)(4000, 0)
            };
            \addlegendentry{\small{KD-based Private CF}}
        
        \addplot[
            color=teal,
            mark=triangle,
            ]
            coordinates {
             (0, 0)(50, 57)(100, 60)(200,62)(300, 64)(500, 66)(1000,67)%(2000, 0)(3000, 0)(4000, 0)
            };
            \addlegendentry{\small{Direct Private CF}}
            
        \addplot[
            color=red,
            mark=square,
            ]
            coordinates {
            (0, 0)(50, 65)(100, 68)(200,71)(300, 74)(500, 75)(1000,80)%(2000, 0)(3000, 0)(4000, 0)
            };
            %\addlegendentry{KD - CF}
        
        \addplot[
            color=blue,
            mark=square,
            ]
            coordinates {
              (0, 0)(50, 60)(100, 64)(200,67)(300, 70)(500, 71)(1000,78)%(2000, 0)(3000, 0)(4000, 0)
            };
            %\addlegendentry{Direct CF}
        
        \addplot[
            color=black,
            mark=square,
            ]
            coordinates {
            (0, 0)(50, 59)(100, 60)(200,65)(300, 66)(500, 68)(1000,75)%(2000,0)(3000, 0)(4000, 0)
            %(0, 0)(50, 64)(100, 74)(200,77)(300, 77)(500,0)(1000,0)(2000, 0)(3000, 0)(4000, 0)
            };
            %\addlegendentry{KD - no CF}
         
        \addplot[
            color=green,
            mark=square,
            ]
            coordinates {
            (0, 0)(50, 57)(100, 58)(200,64)(300, 64)(500, 65)(1000,74)%(2000, 0)(3000, 0)(4000, 0)
            %(0, 0)(50, 63)(100, 73)(200,76)(300, 0)(500,0)(1000,0)(2000, 0)(3000, 0)(4000, 0)
            };
            %\addlegendentry{Direct - no CF}
        \end{axis}
        \end{tikzpicture}
       \caption{Housing Dataset}
       \label{fig:housing}
      \end{subfigure}
  \caption{The agreement values achieved by the MEA approach in the various scenarios across the a) GMSC dataset, b) Credit Fraud Dataset, c) Housing Dataset with respect to number of queries made.}
  \label{fig:all_datasets}
\end{figure*}
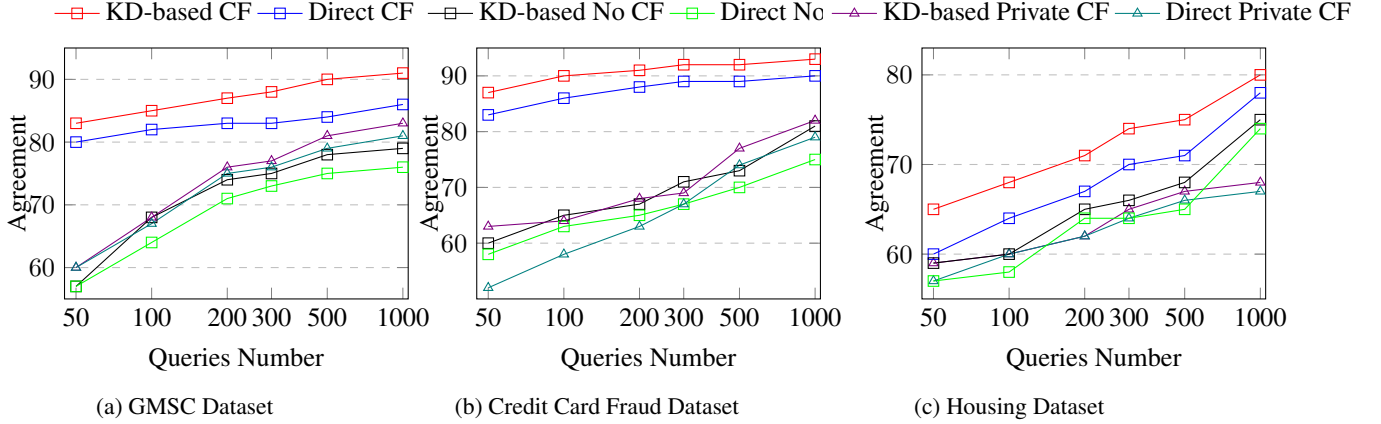
% most important things first - agreement 
    % confirming CF vs. Direct No CF (Impact of CF) (1) also the other methodologies confirm this (KD) --------> OK
    % KD stronger than Direct --------> OK
        % proposed KD with CF vs. Direct with CF --------> OK
        % KD without CF vs. Direct without CF --------> OK
    % Private CF: Quantify how much they protect
        % DP CF vs Traditional CF (KD - Direct)
        % IN DP, KD > Direct (slightly better)
        % Mention with DP CF attacks comparable to no CF
    % something achieves agreement between x and y and another, compare the number of queries.
%We first conduct experiments to confirm the efficacy of CFs in performing MEA and training a threat model compared to scenarios where CFs are not present. As illustrated in Fig. \ref{fig:all_datasets}, we present the agreement after applying our experiments while varying the number of queries.

We start our discussion by analyzing the \emph{agreement} obtained by the employed approaches for MEA in the various scenarios. During the training phase of the threat model, the attacker interacts with the model by providing it with randomly generated data points. In our experimentation, we systematically increase the number of queries to conduct a comprehensive analysis of their impact on MEA.% and their effectiveness in fastening the MEA compared to normal queries.

Figure \ref{fig:all_datasets} shows the \emph{agreement} achieved by \emph{KD-based MEA} and \emph{Direct} in the various scenarios with respect to the number of queries made to the API across the GMSC, the Credit Card Fraud, and the California Housing datasets. Results show a consistent pattern across all scenarios as agreement initially increases significantly as the number of queries increases until a point where additional queries cease to yield a notable increase in agreement. This suggests that there is a saturation point in the MEA, and further queries may not necessarily contribute to enhancing the agreement.

%We observe a trend in which the introduction of additional queries initially leads to a substantial boost in agreement. However, this upward trend eventually

%CF vs. No-CF
\textbf{We now focus on analyzing the effectiveness of exploiting CFs for performing MEAs}, comparing the agreement achieved by each of \emph{KD-based MEA} and \emph{Direct} when CF explanations are used to that when only data points are used (i.e., comparing \emph{Direct No CF} to \emph{Direct-CF}, and \emph{KD-Based No CF} to \emph{KD-based CF}). In both cases, when CFs are exploited, the MEA is more effective, achieving an agreement significantly higher than its counterpart. For instance, in GMSC dataset, \emph{Direct-CF} achieves an agreement ranging between 80 and 85 while \emph{Direct No CF} achieves an agreement ranging between 55 and 75 (showing an additional agreement of up to 25\%). For Credit Card Fraud dataset, the difference is significantly larger as \emph{Direct-CF} shows an agreement ranging between 83 and 90 while Direct-no-CF has an agreement ranging between 58 and 75. This difference is also observed for California Housing dataset where exploiting CFs allows for an additional agreement of around 10\%. Similarly, analyzing the agreement achieved when employing our proposed approach with CFs to its counterpart (i.e., \emph{KD-based CF} vs. \emph{KD-Based No CF}) across the three data sets, we notice that exploiting CFs permits obtaining up to additional 27\% of agreement (for GMSC dataset at 50 queries) and, in the worst case, an additional 4\% on agreement (for Housing dataset at 1000 queries). These results confirm previous findings in literature (e.g., \cite{aivodji2020model} and \cite{wang2022dualcf}) which show the impact of using model's CF explanations for performing MEA.

% KD vs. Direct
\textbf{We now compare our proposed approach (\emph{KD-based MEA}) to \emph{Direct}} in each of the three scenarios (no CF, CF, and Private CF). Results show that in all scenarios, and across all data sets, our proposed approach outperforms \emph{Direct}. In the case of \emph{KD- based No CF} maintains an additional 3\% to 6\% of agreement across all data sets, with respect to \emph{Direct no CF}. In terms of number of queries, \emph{KD-based CF} achieves an agreement of 82\% and 85\% with only 50 and 100 queries, a performance that is only reached by \emph{Direct CF} with x6 number of queries (300) and x10 number of queries (500), respectively. A similar finding can be extracted from the Credit Card Fraud dataset, as KD-CF achieves with 100 queries an agreement of 90\%, which is only attainable by Direct-CF using 1000 queries. Such findings are also attainable by analyzing the performance achieved by these two scenarios with the California Housing data set. 

In the case when CFs are provided by the MLaaS and are then exploited by the attacker, i.e., \emph{KD-based CF} vs. \emph{Direct CF}, results show that across all the 3 datasets, the agreement achieve by \emph{KD-based CF} agreement always surpasses the Direct-CF, independent of the number of queries. In the case \emph{KD-based CF} maintains an additional 2\% to 6\% agreement across all datasets. Similarly, in the case where the MLaaS provides private CFs, i.e., comparing \emph{KD-based Private CF} to \emph{Direct Private CF}, \emph{KD-based Private CF} maintains a slightly additional range from 1\% to 2\% in most query numbers across all datasets. The observed decrease in agreement of MEA with private CFs may be attributed to the privacy-preserving nature of DP. Differential privacy introduces intentional noise into the data to protect individual privacy, making it more challenging for attackers to accurately extract sensitive information. In the context of \emph{KD-based MEA}, when DP is employed, the noise introduced during the training process to achieve privacy guarantees could interfere with the teacher-student model's ability to transfer knowledge much more effectively.

In this consideration, using CFs significantly reduces the number of instances required for MEA to reach a specific high agreement level. For example, on the GMSC dataset, achieving 80\% agreement with CFs took just 50 queries, while accomplishing the same level without CFs required over 1000 queries. This pattern holds across datasets, such as the Credit Card Fraud Dataset, where 83\% agreement needed only 50 CF queries compared to over 1000 queries without CFs. In essence, the use of CFs consistently leads to a substantial reduction in the number of queries needed to achieve high agreement levels across diverse datasets. In addition to that, our proposed methodology KD-based MEA consistently results in higher agreement levels of MEA compared to the baseline approach. This holds in various scenarios and is independent of whether the MLaaS provides users with CF explanations, and across diverse datasets and query numbers. This analysis emphasizes the effectiveness of our proposed method, utilizing KD to achieve superior agreement levels. Importantly, when CFs are employed, MEA requires significantly fewer instances to reach a specific high agreement level. This highlights the efficiency of incorporating CFs in conjunction with KD to enhance the performance of the MEA.

\textbf{We now analyze the impact of generating private CFs through incorporating DP in the explainer}, i.e., by incorporating DP during the generation process of explanations, on mitigating MEA. Specifically, we aim to quantify the extent of protection offered by providing users with private CFs (as opposed to non-private CFs) such as to prevent attackers from leveraging vulnerabilities associated with non-private CFs. To this end, we compare the performance of each of \emph{KD-based CF} and \emph{Direct CF} to its private CF counterparts (\emph{KD-based Private CF} and \emph{Direct Private CF}, respectively). For the GMSC dataset, results show that for the relatively low number of queries (50 to 200), the difference is vast between each scenario with private CF and its counterpart. For instance, \emph{KD-based Private CF} achieves an agreement ranging between 60\% and 75\% while that \emph{KD-based CF} obtains an agreement between 83\% and 87\%.

In the case of the GMSC dataset, we observe that the incorporation of DP shows that our proposed strategy can maintain agreement levels comparable to scenarios without CFs and less than when incorporating traditional CFs, with an agreement range of 60\% to 83\% with \emph{KD-based CF} and 60\% to 81\% with direct training, compared to 57\% to 79\% with \emph{KD-based MEA no CF} and a range of 57\% to 76\% with \emph{Direct No CF}. For the credit card fraud dataset, a similar trend is observed. With \emph{KD-based Private CF}, we achieve a range of 63\% to 82\% and 52\% to 79\% for a direct-DP train, compared to \emph{KD-based no CF} of 60\% to 81\% and a range of 58\% to 75\% with \emph{Direct No CF}. For the California housing dataset, the agreement of \emph{KD-based CF} ranges from 59\% to 68\% with KD and 57\% to 67\% with \emph{Direct Private CF}, compared to 59\% to 75\% with \emph{KD-based no CF}, and direct no CF ranges from 57\% to 74\%. It is also worth noting that, \emph{KD-based Private CF}is slightly better than direct training over all datasets.

Results suggest that the incorporation of DP can play a crucial role in maintaining agreement levels comparable to scenarios without CFs, as they show that across the three datasets a mitigation can be achieved. In the appendix, we add further discussion about the impact of DP on the generation of CF and discuss what could be the potential reasons behind such results.
% while still providing a CF within the API.
\section{Conclusion}
In this paper, our primary focus is on elucidating the privacy implications associated with providing Counterfactual Explanations (CFs) within a Machine Learning as a Service (MLaaS) platform. We first examine how CFs can be exploited for performing effective Model Extraction Attacks (MEA). We propose a novel approach based on Knowledge Distillation (KD) to effectively train a substitute model. Then, we propose to incorporate Differential Privacy (DP) into the CF generator and we assess to what extent it can serve as a mitigation strategy against MEAs, and explore its potential influence on the quality of generated CFs. Experimental results affirm that CFs can be vulnerable to exploitation for MEA. Moreover, results show that our proposed KD-based MEA approach outperforms the baseline approach. The results also demonstrate that integrating DP into the CF generation process effectively mitigates MEA while preserving to an acceptable extent the quality of CF explanations. This work represents a crucial step toward the dual objectives of preserving privacy and maintaining the integrity of explanation quality. For future work, we aim to conduct a more in-depth analysis of the impact of DP on the CF generation process, and its role in mitigating other types of attacks. This will involve a sensitivity analysis, focusing on the impact of noise levels and the guarantees provided by DP in the context of CF generation.

%\section{Acknowldgment}

%We proposed a novel method for conducting MEA, by leveraging the principles of Knowledge Distillation (KD). Additionally, we propose to incorporate Differential Privacy (DP) in the CF generation process, aiming to address MEA mitigation while still delivering meaningful CFs. We conducted extensive evaluation, we compared our proposed methodology with baselines that usually employ direct training of CFs to build the extracted model. Furthermore, we delve into the analysis of the impact of DP on MEA, comparing it with baseline scenarios where CFs are provided or not provided. Our findings affirm that CFs can be vulnerable to exploitation for MEA and that our proposed methodology surpasses direct training with a low number of queries. Additionally, our results demonstrate that integrating DP into the CF generation process effectively mitigates MEA while preserving the quality of CF explanations. This work represents a crucial step toward the dual objectives of preserving privacy and maintaining the integrity of explanation quality. For future work, we aim to conduct a more in-depth analysis of the DP effect on CF generation and its role in mitigating MEA. This will involve a sensitivity analysis, focusing on the impact of noise levels and the guarantees provided by DP in the context of CF generation.

\bibliographystyle{abbrv}

%\newpage
\bibliography{references}
\newpage

\appendix

\section{Appendix / supplemental material}

\subsection{Preliminaries}
In this section, we discuss the key concepts used in our work. We first introduce the concept of CF explanations. Then, we provide an overview of DP. Finally, we introduce the concept of KD.

\subsubsection{Counterfactual Explanations (CF)}
A CF explanation is a type of example-based explanation that provides a hypothetical scenario that illustrates how a different decision or outcome could have arisen if the input data had been altered \cite{wachter2017counterfactual}. CFs provide users with an actionable explanation as they enable users to understand how changes in inputs would affect the model's output \cite{poyiadzi2020face}. Additionally, CFs allow to identification of the variables that should have differed in a given input to observe a different outcome, thus making it possible to assess the influence of specific factors, which in turn provides valuable insights into how decisions are made by an ML model \cite{wachter2017counterfactual}. This type of CF analysis has already proven to improve the interpretability of ML models in several domains, such as healthcare and finance \cite{dovsilovic2018explainable}.
%Consider a case where we have a binary classification: CFs, for example, will explain the minimal change to what features to change and by how much to change a prediction from class 0 to class 1 (or class 1 to class 0).

To generate CFs, various CF explanation methods obtain CFs by optimizing a customized cost function. These explainers employ distinct strategies such as mixed-integer linear optimization \cite{kanamori2021ordered}, heuristic search strategies \cite{martens2014explaining,moore2019explaining}, and metaheuristic approaches \cite{sharma2019certifai,hashemi2020permuteattack, numeroso2020explaining,nguyen2021counterfactual}, such as genetic algorithms \cite{sharma2019certifai,hashemi2020permuteattack}, Reinforcement Learning \cite{ezzeddine2023sac, chen2021relace}, Generative Adversarial networks \cite{nemirovsky2020countergan} or Graph density bases \cite{poyiadzi2020face}. In our work, we employ the method of Generating CFs for Real-Time Recourse and Interpretability using Residual GANs (CounteRGAN) as a CF explainer \cite{nemirovsky2020countergan}. The authors of CounteRGAN formalized a novel residual generative adversarial network that trains the generator to produce residuals that are intuitive to the notion of perturbations used in counterfactual searches. The search process seeks to maximize the value function concerning the discriminator $D$ and minimize it concerning the generator $G$. Where $C_t$ is the target classifier to be explained, Reg(G, {$x_i$}) is a regularization term, and $x_i$ are samples drawn from the entire data distribution \cite{nemirovsky2020countergan}.

\begin{equation}
\begin{split}
V_{\text{CounterGAN-bb}}(D, G) = & \frac{\sum_{i} C_t(x_i) \log D(x_i)}{\sum_{i} C_t(x_i)} \\
& + \frac{1}{N} \sum_{i} \log (1 - D(x_i + G(x_i))) \\
& + \text{Reg}(G, \{x_i\}),
\end{split}
\end{equation}

CF generators aim to find in-distribution points by optimizing metrics of actionability such as sparsity and proximity \cite{mothilal2020explaining}. Sparsity measures how many features of the CF data point are different from the original data point. Proximity measures the overall distance between the CF and the original data point, where a low proximity value indicates that the two data points are similar. This can be useful for providing actionable feedback, as it suggests that the model is recommending only a few changes that are likely to have a significant impact on the output.

\subsubsection{Knowledge Distillation}
Knowledge distillation, also known as model distillation, is a process that involves the transfer of knowledge from a high-complex model, known as a \textit{teacher model}, to a reduced-complexity model, known as a \textit{student model}, that remains operable within real-world constraints \cite{cho2019efficacy}. While KD constitutes a specific instance of model compression, it serves as a means to extract essential insights, patterns, and expertise embedded in the larger models to create a deployable model. The student model follows a training procedure with the primary objective of imitating the performance of the teacher model. This knowledge transfer mechanism involves the student model learning not only the surface-level predictions made by the teacher model but also the deeper patterns, generalizations, and decision-making strategies embedded within the teacher architecture. To achieve this, the student model is guided during training by incorporating two primary sources of information: the actual target labels or predictions for the dataset at hand, and the soft labels or probability distributions generated by the teacher model in response to the same dataset.

KD necessitates the availability of a well-trained teacher model, a trainable student model, and the specification of a student loss function for assessing predictions against ground-truth labels ($L$). A distillation loss function, accompanied by a temperature parameter ($temp$), is employed to bridge the knowledge gap by comparing the soft predictions of the student to the softened teacher labels. The student and the distillation losses are weighted via an alpha ($\alpha$) which is essential for balancing task-specific accuracy and knowledge transfer.
%In the student model training, starting with the forward pass, both the teacher and student models are activated to make predictions.
Subsequently, the losses incurred are computed, incorporating a weighted combination of the student loss (weighted by $\alpha$) and the distillation loss (weighted by $1 - \alpha$). This weighting mechanism allows for a fine-tuned balance between preserving task-specific accuracy (student loss) and incorporating the knowledge distilled from the teacher model (distillation loss).
%Following the loss calculation, a backward pass through the network is executed, facilitating the adjustment of model parameters in pursuit of optimizing the overall loss, a pivotal step in the iterative training process.
The Kullback-Leibler (KL) divergence, denoted as $KL(P \parallel Q)$ (Eq. \ref{KL}), is a mathematical measure employed in KD to assess the difference between two probability distributions, $P$ and $Q$, where the summation is taken over all classes or categories.

Suppose $P$ represents the soft probabilistic predictions of a teacher model, while $Q$ represents the corresponding probabilistic predictions of a student model (expressed as the output made by a softmax activation function). During training, KL divergence as the distillation loss, encourages the student model to capture the nuances and uncertainties present in the teacher's predictions. In summary, the loss to be optimized by the student is shown by Eq. \ref{loss}.
%\begin{equation}
%\label{softmax}
%    \text{softmax}(x)_i = \frac{e^{x_i}}{\sum_{j=1}^{N} e^{x_j}}
%\end{equation}
\begin{equation}
\label{KL}
   \text{$distillation\_loss$} = \text{$KL(P \parallel Q)$} = \sum_{i} P(i) \log\left(\frac{P(i)}{Q(i)}\right)
\end{equation}
%\vspace{-10pt}
%\begin{equation}
%\label{distillationLoss}
%    \text{$distillation\_loss$} = KL(P \parallel Q)
%\end{equation}
%\vspace{-10pt}
\begin{equation}
\label{loss}
    \text{$loss$} = \alpha \cdot \text{$student\_loss$} + (1 - \alpha) \cdot \text{$distillation\_loss$}
\end{equation}

\subsubsection{Differential Privacy}
Differential privacy is a mathematical framework applied to safeguard individual records by introducing controlled noise to the data and allowing the extraction of valuable insights while ensuring that individual identities remain protected \cite{dwork2006calibrating}. This enables analyzing data without disclosing sensitive information about any specific individual in the dataset \cite{dwork2006differential}. DP aims to guarantee, by definition, that the inclusion or exclusion of any individual record in the dataset should have minimal impact on the outcome of the mechanism. A mechanism refers to any mathematical computation applicable to and with the data. In the context of classification tasks, DP analyzes how the output undergoes probabilistic changes based on a given input. Hence, a mechanism guarantees differential privacy if the likelihood of any outcome is nearly identical for any two datasets that vary by only one record.
%A crucial aspect of DP is its ability to offer a privacy assurance that remains effective irrespective of the knowledge or actions of an adversary, i.e., an individual or entity seeking to extract sensitive information about individual records from the data.

One widely utilized approach for addressing numerical inquiries involves the incorporation of randomized approaches that add calibrated random noise, which works by introducing sufficient noise to the input or output of the mechanism to obscure the potential contribution of any individual record in the data while simultaneously maintaining the overall accuracy of the analysis. In Def. \ref{def:differential-privacy}, we detail the DP inequality.
\begin{definition}[Differential Privacy]
A randomized algorithm $M$ with domain $\mathbb{N}^2$ is $(\varepsilon, \delta)$-differentially private if for all $S \subseteq \text{Range}(M)$ and for all $x, y \in \mathbb{N}^2$ such that $||x - y||_1 \leq 1$:
\[
\Pr[M(x) = S] \leq e^{\varepsilon} \cdot \Pr[M(y) \in S] + \delta,
\]
\label{def:differential-privacy}
\end{definition}
\vspace{-10pt}
Def. \ref{def:differential-privacy} states that an algorithm $M$ ($\varepsilon, \delta$) is differentially private if, for all subsets $S$ of the range of $M$ and all pairs of inputs $x$ and $y$ differing by at most one data instance, the probability of $M(x)$ outputting $S$ is bounded by a factor of $e^{\varepsilon}$ and an error margin $\delta$.
%Previously, DP was employed in creating differentially-private synthetic data, the models for synthetic generation learn the distribution of the original data using a differentially private trained algorithm. Once a privacy budget is established, the algorithm can be employed to produce synthetic data. The resultant data will exhibit the characteristic of differential privacy, making it more challenging to deduce individual records in the original dataset from the newly generated data.

In addition to the perturbation of input or output data, it is possible to achieve DP during the training of ML models and safeguard the privacy of the training model itself. This is usually achieved by introducing perturbations to the model weights and gradients during training \cite{abadi2016deep, zhang2012functional}. The process proceeds as follows: consider an ML model, to be trained on dataset $D$, with a parameter set $w^*$ will minimize an objective $L_D(w) = \sum_{t_i \in D} L(t_i, w)$, DP is injected during the optimization process to prevent the model from memorizing specific details about individual data points. DP training involves multiple steps of adding noise to the gradient of the model parameters concerning the training data being trained on, followed by gradient clipping. Then, the resulting parameter set $w^*$, minimizes the loss that can be later derived.

\subsection{Datasets}
We perform our experiments using three classification datasets, namely, Give Me Some Credits dataset \cite{GMSC}, Credit Card Fraud Dataset \cite{CreditFraud}, and California Housing Dataset \cite{CalifHousing}. The datasets are described as follows.
\begin{itemize}
    \item \textbf{Give Me Some Credits \cite{GMSC}} This dataset is collected to forecast the likelihood of an individual undergoing financial distress within the next two years based on their financial and demographic features. The full dataset encompasses 150,000 applicants, with 139,974 applicants classified as good and 10,026 applicants labeled as bad. We follow the pre-processing of \cite{GMSCDL}.

    \item \textbf{Credit Card Fraud \cite{CreditFraud}} The dataset contains transactions conducted by European cardholders using credit cards in September 2013. Within this dataset, transactions recorded over two days reveal the occurrence of 492 fraudulent instances out of a total of 284,807 transactions.

    \item \textbf{California Housing \cite{CalifHousing}} A dataset was created based on the 1990 U.S. census, consisting of 8 features and a target variable representing the median house value for California districts in dollars. The target variable is then converted into two classes using a threshold defined by the median.
\end{itemize}

\subsection{Impact of Incorporating DP in CF generator on quality of explanations}

We shift our focus to analyzing the impact of incorporating DP within the CounteRGAN CF generator on the quality of CFs\footnote{Note that this analysis is related to the explainer and the quality of the generated CFs, not to the methodology adopted to perform MEA.} in terms of the metrics introduced in Sec. \ref{EvaluationMetrics}.

\begin{table*}[t]
\centering
\caption{Actionability and Prediction Gain of CFs generated on the random set with and without the adoption of DP in the generator.}
\begin{tabular}{c|cc|cc}
           & \multicolumn{2}{|c}{Prediction Gain}              & \multicolumn{2}{|c}{Actionability} \\ \hline
Data       & CFs     & Private CFs                      & CFs     & Private CFs      \\ \hline
GMSC       & 0.243 $\pm$ 0.011 & 0.121 $\pm$ 0.01       & 24.567 $\pm$ 0.364 & 16.981 $\pm$ 0.158 \\
Credit Fraud & 0.700 $\pm$ 0.084 & 0.445 $\pm$ 0.06       & 35.269 $\pm$ 0.328 & 10.507 $\pm$ 0.238 \\
Housing    & 0.633 $\pm$ 0.052 & 0.678 $\pm$ 0.024      & 3.852 $\pm$ 0.053  & 1.004 $\pm$ 0.016
\end{tabular}
\label{othermetrics}
\end{table*}
%We now move into analyzing the impact on the CF generation process after incorporating DP in CounteRGAN explainer and on the CF themselves in terms of the metrics introduced in Sec. \ref{EvaluationMetrics}. 
Table \ref{othermetrics} reports the prediction gain of CounteRGAN and the actionability of the CFs achieved with and without DP across the three datasets. Note that the focus lies on comparing the metrics on the CFs exclusively between scenarios with and without DP, regardless of the applied MEA approach.
Results show that CounteRGAN achieves a higher prediction gain (a higher probability shift, desired) when DP is not incorporated within the explainer than when employed for GMSC (0.243 $\pm$ 0.011 versus 0.121 $\pm$ 0.01) and Credit Fraud datasets (0.700 $\pm$ 0.084 versus 0.445 $\pm$ 0.06) while for Housing dataset, CounteRGAN shows a prediction gain slightly higher in the case where DP is incorporated rather than when not (0.678 versus 0.633). These findings highlight that the integration of DP has an impact on prediction gain, constraining the explainer's progress toward the desired class.

In terms of actionability which is measured as the degree of perturbation or modification of the CF compared to the initial point, the results show that the explainer generates CFs with lower values of actionability when DP is incorporated than when not across all datasets. For instance, for GMSC dataset, the actionability of CFs with DP is 16.981 ($\pm$ 0.158) while that without DP is 24.5 ($\pm$ 0.364). Similarly, for credit fraud and housing datasets, the actionability is reduced when incorporating DP from 35.269 $\pm$ 0.328 to 10.507  $\pm$ 0.238, and from 3.852 $\pm$ 0.053 to 1.004 $\pm$ 0.016 respectively. This means that that the practical usefulness of the CFs or perturbations in guiding actionable decisions is reduced, suggesting that the privacy constraints imposed might compromise the effectiveness of CF analyses in providing actionable guidance for influencing desired outcomes.

Although a higher prediction gain and lower actionability are preferable when generating CFs, the results suggest that the CF generation with DP has taken a different trajectory than it has taken without DP, leading to a reduction in prediction gain. This shift in the CF generation approach is reflected in the probability, ultimately contributing to the observed decrease in agreement, thereby motivating the smaller agreement for MEA.

\begin{table*}[t]
\centering
\caption{Realism of the random points and the CFs generated with and without incorporating DP in the generator across the datasets.}
\begin{tabular}{c|ccc}
           & \multicolumn{3}{c}{Realism}                                   \\ \hline
Data       & random points      & CFs      & Private  CFs      \\ \hline
GMSC       & 15.649 $\pm$ 0.033 & 8.56 $\pm$ 0.142   & 15.723 $\pm$ 0.033  \\
Credit Fraud & 3.104 $\pm$ 0.019  & 3.072 $\pm$ 0.1647 & 4.73 $\pm$ 0.027    \\
Housing    & 2.070 $\pm$ 0.04   & 1.356 $\pm$ 0.031  & 2.0 $\pm$ 0.01 
\end{tabular}
\label{realism}
\end{table*}
To investigate the impact of incorporating DP in the explainer in more detail, we analyze the realism (which allows us to quantify how well a data instance fits a data distribution of a dataset) of the data points, including CFs in the scenarios (no CF, CF, Private CF).

Table \ref{realism} reports the realism of the data points used as initial queries and the CFs generated by the explainer across the three data sets. Results show that the Realism of CFs without incorporating DP in the generator is the lowest in comparison to the Realism of random points and that of CFs when DP is employed, meaning that the CF generator has produced more realistic data points (lower Realism), attempting to align the CFs of random points with the distribution of the training data. In particular, for the GMSC dataset, the randomly generated points have an average realism of 15.6 ($\pm$ 0.03), and the CF generator has produced corresponding CFs with a realism of 8.5 ($\pm$ 0.14). Similarly, for the Credit Fraud data set, the Realism of random points is 3.104 ($\pm$ 0.019) while for the corresponding CFs, it is slightly lower at 3.07 ($\pm$ 0.16), respectively. For the Housing data set, the realism of data points is 2.07 ($\pm$ 0.04) while their corresponding CFs had a realism of 1.35 ($\pm$ 0.03). This implies that the explainer generates CFs with lower realism from the original queried random points and aligns them more closely with the distribution of the training data.
The private CFs preserve a realism very similar to that of random points. More specifically, for the GMSC dataset, the private CFs have preserved their initial average level of realism of 15.723 ($\pm$ 0.033). Similarly, the housing dataset preserves a realism of 2.0 ($\pm$ 0.01). For the Credit fraud dataset, the realism of the CFs deviates by one degree from the distribution, resulting in a realism of 4.73 ($\pm$ 4.73). With the incorporation of DP, realism can be preserved indicating the effectiveness of the Private CFs in maintaining the quality and distribution of the original data. The findings might vary depending on the dataset, as shown by the slight deviation in realism for the Credit Fraud dataset.

It is worth noting that, random points are inherently unrealistic (do not exhibit high realism). Our private CF generation approach ensures this unrealistic nature is preserved when queried with a random data point. This is crucial because our experiments showed that traditional CFs, without privacy protections, generate more realistic outputs from random data, potentially revealing private information.

\subsection{Limitations and Future Work}
The limitations of our current approach and potential avenues for future research are the following:
\begin{itemize}
    \item \textbf{Attack Scope}: Our work focuses on MEA leveraging CFs, it is important to acknowledge the potential existence of other privacy attacks such as membership inference and model inversion that warrant further investigation. To this end, in future work, we plan to evaluate the effectiveness of our proposed private GAN-based CF generation approach against other attack types.

    \item \textbf{Privacy-Performance Trade-off}: In this work, we made the first attempt to integrate DP in GAN-based CF generators. We assume that GAN-based CFs should be privacy-preserving against random and zero-knowledge queries. This means it should not reveal any realistic CFs when queried by random noise. A future direction is to thoroughly evaluate the impact of various DP levels on considering both their influence on mitigating various privacy attacks including MEA and their impact on the effectiveness and quality of CFs with DP.

    \item \textbf{Focus on Deep Learning Applications}: KD is effective with DNNs due to its complex architecture and ability to learn rich data representations. Future research aims to explore adapting KD for other ML algorithms such as tree and ensemble-based algorithms and to potentially unlock similar performance enhancements.
\end{itemize}

\section*{NeurIPS Paper Checklist}
\begin{enumerate}

\item {\bf Claims}
    \item[] Question: Do the main claims made in the abstract and introduction accurately reflect the paper's contributions and scope?
    \item[] Answer: \answerYes{} % Replace by \answerYes{}, \answerNo{}, or \answerNA{}.
    \item[] Justification: We summarized our contributions in the abstract and introduction.
    % \item[] Guidelines:
    % \begin{itemize}
    %     \item The answer NA means that the abstract and introduction do not include the claims made in the paper.
    %     \item The abstract and/or introduction should clearly state the claims made, including the contributions made in the paper and important assumptions and limitations. A No or NA answer to this question will not be perceived well by the reviewers. 
    %     \item The claims made should match theoretical and experimental results, and reflect how much the results can be expected to generalize to other settings. 
    %     \item It is fine to include aspirational goals as motivation as long as it is clear that these goals are not attained by the paper. 
    % \end{itemize}

\item {\bf Limitations}
    \item[] Question: Does the paper discuss the limitations of the work performed by the authors?
    \item[] Answer: \answerYes{} % Replace by \answerYes{}, \answerNo{}, or \answerNA{}.
    \item[] Justification: The Limitations are discussed as a separate section in the appendix.

\item {\bf Theory Assumptions and Proofs}
    \item[] Question: For each theoretical result, does the paper provide the full set of assumptions and a complete (and correct) proof?
    \item[] Answer: \answerNA{} % Replace by \answerYes{}, \answerNo{}, or \answerNA{}.
    \item[] Justification: All assumptions stated are referenced.
    % \item[] Guidelines:
    % \begin{itemize}
    %     \item The answer NA means that the paper does not include theoretical results. 
    %     \item All the theorems, formulas, and proofs in the paper should be numbered and cross-referenced.
    %     \item All assumptions should be clearly stated or referenced in the statement of any theorems.
    %     \item The proofs can either appear in the main paper or the supplemental material, but if they appear in the supplemental material, the authors are encouraged to provide a short proof sketch to provide intuition. 
    %     \item Inversely, any informal proof provided in the core of the paper should be complemented by formal proofs provided in appendix or supplemental material.
    %     \item Theorems and Lemmas that the proof relies upon should be properly referenced. 
    % \end{itemize}

    \item {\bf Experimental Result Reproducibility}
    \item[] Question: Does the paper fully disclose all the information needed to reproduce the main experimental results of the paper to the extent that it affects the main claims and/or conclusions of the paper (regardless of whether the code and data are provided or not)?
    \item[] Answer: \answerYes{} % Replace by \answerYes{}, \answerNo{}, or \answerNA{}.
    \item[] Justification: All the steps required for reproducibility are provided in the experimental settings.

\item {\bf Open access to data and code}
    \item[] Question: Does the paper provide open access to the data and code, with sufficient instructions to faithfully reproduce the main experimental results, as described in supplemental material?
    \item[] Answer: \answerYes{} % Replace by \answerYes{}, \answerNo{}, or \answerNA{}.
    \item[] Justification: We use publicly available datasets (references to datasets are reported in the paper). We provide sufficient instructions and code to reproduce the main experimental results (libraries, parameters, etc.).
    % \item[] Guidelines:
    % \begin{itemize}
    %     \item The answer NA means that paper does not include experiments requiring code.
    %     \item Please see the NeurIPS code and data submission guidelines (\url{https://nips.cc/public/guides/CodeSubmissionPolicy}) for more details.
    %     \item While we encourage the release of code and data, we understand that this might not be possible, so “No” is an acceptable answer. Papers cannot be rejected simply for not including code, unless this is central to the contribution (e.g., for a new open-source benchmark).
    %     \item The instructions should contain the exact command and environment needed to run to reproduce the results. See the NeurIPS code and data submission guidelines (\url{https://nips.cc/public/guides/CodeSubmissionPolicy}) for more details.
    %     \item The authors should provide instructions on data access and preparation, including how to access the raw data, preprocessed data, intermediate data, and generated data, etc.
    %     \item The authors should provide scripts to reproduce all experimental results for the new proposed method and baselines. If only a subset of experiments are reproducible, they should state which ones are omitted from the script and why.
    %     \item At submission time, to preserve anonymity, the authors should release anonymized versions (if applicable).
    %     \item Providing as much information as possible in supplemental material (appended to the paper) is recommended, but including URLs to data and code is permitted.
    % \end{itemize}

\item {\bf Experimental Setting/Details}
    \item[] Question: Does the paper specify all the training and test details (e.g., data splits, hyperparameters, how they were chosen, type of optimizer, etc.) necessary to understand the results?
    \item[] Answer: \answerYes{} % Replace by \answerYes{}, \answerNo{}, or \answerNA{}.
    \item[] Justification: We provide all the necessary informations.
    % \item[] Guidelines:
    % \begin{itemize}
    %     \item The answer NA means that the paper does not include experiments.
    %     \item The experimental setting should be presented in the core of the paper to a level of detail that is necessary to appreciate the results and make sense of them.
    %     \item The full details can be provided either with the code, in appendix, or as supplemental material.
    % \end{itemize}

\item {\bf Experiment Statistical Significance}
    \item[] Question: Does the paper report error bars suitably and correctly defined or other appropriate information about the statistical significance of the experiments?
    \item[] Answer: \answerYes{} % Replace by \answerYes{}, \answerNo{}, or \answerNA{}.
    \item[] Justification:  We report the results of the average of 10 runs with randomly chosen subsets for each experiment of the MEA KD-based attack.
    % \item[] Guidelines:
    % \begin{itemize}
    %     \item The answer NA means that the paper does not include experiments.
    %     \item The authors should answer "Yes" if the results are accompanied by error bars, confidence intervals, or statistical significance tests, at least for the experiments that support the main claims of the paper.
    %     \item The factors of variability that the error bars are capturing should be clearly stated (for example, train/test split, initialization, random drawing of some parameter, or overall run with given experimental conditions).
    %     \item The method for calculating the error bars should be explained (closed form formula, call to a library function, bootstrap, etc.)
    %     \item The assumptions made should be given (e.g., Normally distributed errors).
    %     \item It should be clear whether the error bar is the standard deviation or the standard error of the mean.
    %     \item It is OK to report 1-sigma error bars, but one should state it. The authors should preferably report a 2-sigma error bar than state that they have a 96\% CI, if the hypothesis of Normality of errors is not verified.
    %     \item For asymmetric distributions, the authors should be careful not to show in tables or figures symmetric error bars that would yield results that are out of range (e.g. negative error rates).
    %     \item If error bars are reported in tables or plots, The authors should explain in the text how they were calculated and reference the corresponding figures or tables in the text.
    % \end{itemize}

\item {\bf Experiments Compute Resources}
    \item[] Question: For each experiment, does the paper provide sufficient information on the computer resources (type of compute workers, memory, time of execution) needed to reproduce the experiments?
    \item[] Answer: \answerYes{} % Replace by \answerYes{}, \answerNo{}, or \answerNA{}.
    \item[] Justification: Experiments were computed with a machine of intel Core i7, a GPU of RTX 3070, and 8 GB of RAM.
    % \item[] Guidelines:
    % \begin{itemize}
    %     \item The answer NA means that the paper does not include experiments.
    %     \item The paper should indicate the type of compute workers CPU or GPU, internal cluster, or cloud provider, including relevant memory and storage.
    %     \item The paper should provide the amount of compute required for each of the individual experimental runs as well as estimate the total compute. 
    %     \item The paper should disclose whether the full research project required more compute than the experiments reported in the paper (e.g., preliminary or failed experiments that didn't make it into the paper). 
    % \end{itemize}
    
\item {\bf Code Of Ethics}
    \item[] Question: Does the research conducted in the paper conform, in every respect, with the NeurIPS Code of Ethics \url{https://neurips.cc/public/EthicsGuidelines}?
    \item[] Answer: \answerYes{} % Replace by \answerYes{}, \answerNo{}, or \answerNA{}.
    \item[] Justification: We follow the ethics guideline.
    % \item[] Guidelines:
    % \begin{itemize}
    %     \item The answer NA means that the authors have not reviewed the NeurIPS Code of Ethics.
    %     \item If the authors answer No, they should explain the special circumstances that require a deviation from the Code of Ethics.
    %     \item The authors should make sure to preserve anonymity (e.g., if there is a special consideration due to laws or regulations in their jurisdiction).
    % \end{itemize}

\item {\bf Broader Impacts}
    \item[] Question: Does the paper discuss both potential positive societal impacts and negative societal impacts of the work performed?
    \item[] Answer:  \answerNA{} % Replace by \answerYes{}, \answerNo{}, or \answerNA{}.
    \item[] Justification: We discuss the effect of performing a model extraction attack, and we propose a mitigation method.

\item {\bf Safeguards}
    \item[] Question: Does the paper describe safeguards that have been put in place for responsible release of data or models that have a high risk for misuse (e.g., pretrained language models, image generators, or scraped datasets)?
    \item[] Answer: \answerNA{} % Replace by \answerYes{}, \answerNo{}, or \answerNA{}.
    \item[] Justification: The paper poses no such risks.
    %\item[] Guidelines:
    % \begin{itemize}
    %     \item The answer NA means that the paper poses no such risks.
    %     \item Released models that have a high risk for misuse or dual-use should be released with necessary safeguards to allow for controlled use of the model, for example by requiring that users adhere to usage guidelines or restrictions to access the model or implementing safety filters. 
    %     \item Datasets that have been scraped from the Internet could pose safety risks. The authors should describe how they avoided releasing unsafe images.
    %     \item We recognize that providing effective safeguards is challenging, and many papers do not require this, but we encourage authors to take this into account and make a best faith effort.
    % \end{itemize}

\item {\bf Licenses for existing assets}
    \item[] Question: Are the creators or original owners of assets (e.g., code, data, models), used in the paper, properly credited and are the license and terms of use explicitly mentioned and properly respected?
    \item[] Answer: \answerYes{} % Replace by \answerYes{}, \answerNo{}, or \answerNA{}.
    \item[] Justification: We use publicly available datasets and codes from Kaggle that we referenced.
    % \item[] Guidelines:
    % \begin{itemize}
    %     \item The answer NA means that the paper does not use existing assets.
    %     \item The authors should cite the original paper that produced the code package or dataset.
    %     \item The authors should state which version of the asset is used and, if possible, include a URL.
    %     \item The name of the license (e.g., CC-BY 4.0) should be included for each asset.
    %     \item For scraped data from a particular source (e.g., website), the copyright and terms of service of that source should be provided.
    %     \item If assets are released, the license, copyright information, and terms of use in the package should be provided. For popular datasets, \url{paperswithcode.com/datasets} has curated licenses for some datasets. Their licensing guide can help determine the license of a dataset.
    %     \item For existing datasets that are re-packaged, both the original license and the license of the derived asset (if it has changed) should be provided.
    %     \item If this information is not available online, the authors are encouraged to reach out to the asset's creators.
    % \end{itemize}

\item {\bf New Assets}
    \item[] Question: Are new assets introduced in the paper well documented and is the documentation provided alongside the assets?
    \item[] Answer: \answerNA{} % Replace by \answerYes{}, \answerNo{}, or \answerNA{}.
    \item[] Justification: We do not provide new assets.
    % \item[] Guidelines:
    % \begin{itemize}
    %     \item The answer NA means that the paper does not release new assets.
    %     \item Researchers should communicate the details of the dataset/code/model as part of their submissions via structured templates. This includes details about training, license, limitations, etc. 
    %     \item The paper should discuss whether and how consent was obtained from people whose asset is used.
    %     \item At submission time, remember to anonymize your assets (if applicable). You can either create an anonymized URL or include an anonymized zip file.
    % \end{itemize}

\item {\bf Crowdsourcing and Research with Human Subjects}
    \item[] Question: For crowdsourcing experiments and research with human subjects, does the paper include the full text of instructions given to participants and screenshots, if applicable, as well as details about compensation (if any)? 
    \item[] Answer: \answerNA{} % Replace by \answerYes{}, \answerNo{}, or \answerNA{}.
    \item[] Justification: We do not involve crowdsourcing nor research with human subjects.
    % \item[] Guidelines:
    % \begin{itemize}
    %     \item The answer NA means that the paper does not involve crowdsourcing nor research with human subjects.
    %     \item Including this information in the supplemental material is fine, but if the main contribution of the paper involves human subjects, then as much detail as possible should be included in the main paper. 
    %     \item According to the NeurIPS Code of Ethics, workers involved in data collection, curation, or other labor should be paid at least the minimum wage in the country of the data collector. 
    % \end{itemize}

\item {\bf Institutional Review Board (IRB) Approvals or Equivalent for Research with Human Subjects}
    \item[] Question: Does the paper describe potential risks incurred by study participants, whether such risks were disclosed to the subjects, and whether Institutional Review Board (IRB) approvals (or an equivalent approval/review based on the requirements of your country or institution) were obtained?
    \item[] Answer: \answerNA{} % Replace by \answerYes{}, \answerNo{}, or \answerNA{}.
    \item[] Justification: The paper does not involve crowdsourcing nor research with human subjects.
    % \item[] Guidelines:
    % \begin{itemize}
    %     \item The answer NA means that the paper does not involve crowdsourcing nor research with human subjects.
    %     \item Depending on the country in which research is conducted, IRB approval (or equivalent) may be required for any human subjects research. If you obtained IRB approval, you should clearly state this in the paper. 
    %     \item We recognize that the procedures for this may vary significantly between institutions and locations, and we expect authors to adhere to the NeurIPS Code of Ethics and the guidelines for their institution. 
    %     \item For initial submissions, do not include any information that would break anonymity (if applicable), such as the institution conducting the review.
    % \end{itemize}

\end{enumerate}

\end{document}